\documentclass[11pt,a4paper]{article}
\PassOptionsToPackage{breaklinks}{hyperref}
\usepackage[hyperref]{naaclhlt2019}
\usepackage{times}
\usepackage{latexsym}

\usepackage{booktabs}
\usepackage{tabularx}
\usepackage{multirow}
\newcommand{\ra}[1]{\renewcommand{\arraystretch}{#1}}
\usepackage{colortbl}
\definecolor{EsCol}{hsb}{0.05,0.0,1.0}
\definecolor{EnCol}{hsb}{0.6,0.0,1.0}
\definecolor{DeCol}{hsb}{0.3,0.0,1.0}
\definecolor{FrCol}{hsb}{0.9,0.0,1.0}

\definecolor{NACol}{rgb}{0.4,0.4,0.4}

\usepackage{amssymb}

\usepackage{graphicx}

\usepackage{tablefootnote}

\usepackage[utf8]{inputenc}

\aclfinalcopy % Uncomment this line for the final submission
\def\aclpaperid{2424} %  Enter the acl Paper ID here

\title{Strong Baselines for Complex Word Identification\\ across Multiple Languages}

\author{Pierre Finnimore$^1$, Elisabeth Fritzsch$^1$, Daniel King$^1$, Alison Sneyd$^1$,\\ {\bf Aneeq Ur Rehman$^1$}, {\bf Fernando Alva-Manchego$^1$} \and {\bf Andreas Vlachos$^2$}\\
  $^1$Department of Computer Science, University of Sheffield \\
  $^2$Department of Computer Science and Technology, University of Cambridge \\
  {\tt \{pmfinnimore,fritzsch.elisabeth,danielking1903\}@gmail.com,} \\ 
  {\tt a.sneyd@shef.ac.uk,a.neeq8394@gmail.com,f.alva@shef.ac.uk,} \\
  {\tt andreas.vlachos@cst.cam.ac.uk}}

\date{}

\begin{document}
\maketitle
\begin{abstract} % Fernando
Complex Word Identification (CWI) is the task of identifying which words or phrases in a sentence are difficult to understand by a target audience. The latest CWI Shared Task released data for two settings: monolingual (i.e.\ train and test in the same language) and cross-lingual (i.e.\ test in a language not seen during training). The best monolingual models relied on language-dependent features, which do not generalise in the cross-lingual setting, while the best cross-lingual model used neural networks with multi-task learning. In this paper, we present monolingual and cross-lingual CWI models that perform as well as (or better than) most models submitted to the latest CWI Shared Task. We show that carefully selected features and simple learning models can achieve state-of-the-art performance, and result in strong baselines for future development in this area. Finally, we discuss how inconsistencies in the annotation of the data can explain some of the results obtained.
\end{abstract}

\section{Introduction} % Fernando
%\nando{6 - Explain CWI directly and not LS. Work on the first two sentences to make the definition of the task clearer}
%Lexical Simplification (LS) consists of identifying complex words in a text, and then replacing them with simpler synonyms~\citep{survey:shardlow:2014}. 
Complex Word Identification (CWI) consists of deciding which words (or phrases) in a text could be difficult to understand by a specific type of reader.
In this work, we follow the CWI Shared Tasks~\citep{sharedtask:paetzold-specia:2016,sharedtask:yimam-etal:2018} and assume that a target word or multi-word expression (MWE\footnote{We consider n-grams with $n\geq2$ as MWEs, while \citet{sharedtask:yimam-etal:2018} used $n\geq3$.}) in a sentence is given, and our goal is to determine if it is complex or not (an example is shown in Table~\ref{tab:example}).
Under this setting, CWI is normally treated using supervised learning and feature engineering to build monolingual models~\citep{sharedtask:paetzold-specia:2016,sharedtask:yimam-etal:2018}.
Unfortunately, this approach is infeasible for languages with scarce resources of annotated data. 
In this paper, we are interested in both monolingual and cross-lingual CWI; in the latter, we build models to make predictions for languages not seen during training.

\begin{table}[tb]
\centering \small
\ra{1.2}
\begin{tabular}{@{}llc@{}}
\toprule
Sentence                                & Target word/MWE & Complex?\\ 
\midrule
\multirow{3}{0.4\columnwidth}{\it Both China and the Philippines flexed their muscles on Wednesday.}                             & flexed    & Yes\\
                                        & flexed their muscles & Yes\\
                                        & muscles   & No\\
\bottomrule
\end{tabular}
\caption{An annotated sentence in the English dataset of the Second CWI Shared Task.} 
\label{tab:example}
\end{table}

While monolingual CWI has been studied extensively (see a survey in \citet{survey:paetzold-specia:2017}), the cross-lingual setup of the task was introduced only recently by \citet{multicrossdata::yimam-etal:2017}, who collected human annotations from native and non-native speakers of Spanish and German, and integrated them with similar data previously produced for three English domains \citep{cwig2g3:yimam-etal:2017}: News, WikiNews and Wikipedia. 
  %Using language-independent features such as length, frequency, syntax, multilingual word embeddings and topic-relatedness, the authors observed only modest decreases in performance when using cross-lingual models. In one case, there was even an improved result when training on one of the English datasets and testing on German.\nando{a bit odd, not sure we need this at least in this way} 
%AV: Do you think this should happen here? Might be a bit too soon to talk about our models.
%\nando{7 - Provide an example of the data, and use it to explain the task and the terminology we use in the paper.}

For the Second CWI Shared Task~\citep{sharedtask:yimam-etal:2018}, participants built monolingual models using the datasets previously described, and also tested their cross-lingual capabilities on newly collected French data. 
In the monolingual track, the best systems for English \citep{cwi:gooding-kochmar:2018} differed significantly in terms of feature set size and the model's complexity, to the best systems for German and Spanish ~\citep{cwi:kajiwara-komachi:2018}. %\av{What was the difference?}\anote{added explanation}
The latter used  Random Forests with eight features, whilst the former used AdaBoost with 5000 estimators or ensemble voting combining AdaBoost and Random Forest classifiers, with about 20 features. 

In the cross-lingual track, only two teams achieved better scores than the baseline: \citet{cwi:kajiwara-komachi:2018} who used length and frequency based features with Random Forests, and \citet{multitask:bingel-bjerva:2018} who implemented an ensemble of Random Forests and feed-forward neural networks in a multi-task learning architecture. 

Our approach to CWI differs from previous work in that we begin by building competitive monolingual models, but using the same set of features and learning algorithm across languages.
This reduces the possibility of getting high scores due to modelling annotation artifacts present in the dataset of one language. 
Our monolingual models achieve better scores for Spanish and German than the best systems in the Second CWI Shared Task. 
After that, we focus on language-independent features, and keep those that achieve good performance in cross-lingual experiments across all possible combinations of languages. 
This results in a small set of five language-independent features, which achieve a score as high as the top models in the French test set. 
%\av{I would mention some numbers here from the results, e.g. how good our monolingual model is on average, or how good our crosslingual model is when compared to mono/cross-lingual SotAs}
Finally, we analyse the annotation of the datasets and find some inconsistencies that could explain some of our results. 
%\nando{11 - Ensure consistency in the terminology}

%\citet{sharedtask:paetzold-specia:2016} reports the results of the First CWI Shared Task from SemEval 2016. The task introduced a dataset from Simple English Wikipedia, annotated by non-native speakers. The system with the best F-score employed a threshold-based approach to a word's document frequency in Simple English Wikipedia. The system with the best G-score (newly introduced as the harmonic mean of Accuracy and Recall) used System Voting techniques and 69 features, including lexical, semantic and morphological features.

% \citet{multitask:bingel-bjerva:2018} used a multi-task approach for the multilingual CWI task. The authors constructed ensembles of random forests and feed-forward neural  networks which had a multi-task learning configuration. The neural models shared a hidden representation across languages, connected to a language-specific input layer for each language. Using largely language-agnostic features of word log-probability (based on frequencies from Wikipedia dumps), character perplexity, number of synsets, hypernym chain length, inflectional complexity, surface features, parts-of-speech and target-sentence similarity, this system achieved the highest scores on the French cross-lingual track of the Second CWI Shared Task.

Code for all our models can be found at: \mbox{\small \url{https://github.com/sheffieldnlp/cwi}} %\nando{Would it be better to put this link at the end of the Introduction?}\anote{moved}

\section{Problem Formulation}
\label{sec:problem}

We tackle the binary classification task in the Second CWI Shared Task~\citep{sharedtask:yimam-etal:2018}, in which a model decides if a target word/MWE in a sentence is complex or not. 
Following common practice, we extract features from the target word/MWE and its context, and then use a supervised learning algorithm to train a classifier. For training and testing our models, we use the annotated datasets provided for the Second CWI Shared Task (see Table~\ref{tab:dataset_size} for some statistics).
 
\begin{table}[htb]
\centering \small
\begin{tabular}{@{}lrrr@{}}
\toprule
Dataset             & Train             & Dev           & Test  \\
\midrule
English (EN) - News         & 14,002            & 1,764         & 2,095 \\
English (EN) - WikiNews     & 7,746             & 870           & 1,287 \\
English (EN) - Wikipedia    & 5,551             & 694           & 870   \\
Spanish (ES)                & 13,750            & 1,622         & 2,232 \\
German (DE)                 & 6,151             & 795           & 959   \\
French (FR)                 & {\color{NACol}N/A}& {\color{NACol}N/A} & 2,251 \\
\bottomrule
\end{tabular}
\caption{Number of annotated samples in each dataset for each language. \label{tab:dataset_size}}
\end{table}

\section{Monolingual Models}
\label{sec:monolingual}

\subsection{Features Description}
% Dan, Pierre
\label{sec:mono_features}
Our feature set consists of 25 features that can be extracted for all languages considered (English, German, Spanish and French).
%AV: no need
%, and they are based on previous work and morphological intuitions. %\pmfin{Our features, Other people's features, General NLP features?}
%AV: I prefer the division according to their type. I would mention the previous information as we go along.
%
%AV: Group the features and then talk about some of them in particular, especially those that matter to the performance later in the results. Focus on the intuition why they should work
%
They can be divided into three broad categories: features based on the target word/MWE, sub-word level features, and sentence-level features to capture information from the target's context. 
As we intended that our features be applicable across languages, we drew on features found to be useful in previous work on CWI  \citep{multicrossdata::yimam-etal:2017, sharedtask:yimam-etal:2018}. 
We made use of the python libraries spaCy\footnote{\url{https://spacy.io/}} \citep{spacy2} and NLTK\footnote{\url{https://www.nltk.org/}} \citep{nltk}. %\anote{added spacy/nltk citations- is this the correct spacy citation?}
Details on the resources used for extracting each feature can be found in Appendix~\ref{appe:detailed_features}.
%\anote{9: cite features} 
 % in Table \ref{table:mono_features}.
%\pmfin{Three paragraphs: One for each morphological category}
%AV: good! Maybe start with the word-level ones as they are the most obvious starting point??

% 5 future features: len_tokens, num_complex_punct, len_syllables, sent_length, unigram_prob
% len_tokens - word level (number of words in target)
% num_complex_punct - sub-word level (number of dashes or commas or semi-colons as part of the target)
% len_syllables - sub-word level (number of syllables)
% sent_length - sentence context level (number of tokens in sentence)
% unigram_prob - word level

At the \textbf{target word/MWE level}, we experimented with features such as Named Entity (NE) type, part-of-speech, hypernym counts, number of tokens in the target, language-normalised number of characters in each word, and simple unigram probabilities. 
These features are linguistically motivated.
%The perceived complexity of a list of several words is likely to be higher than that of a single word, as each component word can be complex, or simple component words can be synthesised into a complex whole.
The perceived complexity of a MWE may be higher than that of a single word, as each component word can be complex, or simple component words can be synthesised into a complex whole. %\av{I don't get this is a list of several words something different from MWE?}\anote{I didn't write this sentence originally, but I think list of words refers to MWE and I've rewritten it accordingly.}  
Similarly, infrequent words are less familiar, so we would expect low-probability target words to be found more complex. 
Along these lines, proper nouns could be more complex, as there is a vast number of NEs, and the chance that a person has encountered any one of them is low. 
We would expect this trend to reverse for the NE type of organisations, in combination with the Enlgish-News dataset, as organisations mentioned in news articles are frequently global, and so the chance that a person has encountered a proper noun \emph{that is an organisation} is often higher than for proper nouns in general. 
In total, 14 features were used at the target word/MWE level.

% in English, words which were capitalised, as well as words identified as proper nouns were rated more complex, while if the target was a noun phrase, or an organisation, it was rated as simpler. Mirroring the point about sub-words, proper nouns are often difficult to break down, so whether a reader regards the word as complex or not is likely to depend on whether they have encountered it before. Furthermore, organisations mentioned in News articles are frequently global, and so the chance that a person has encountered a proper noun \textit{that is an organisation} is often higher than for proper nouns in general.

Our \textbf{sub-word level} features include prefixes, suffixes, the number of syllables, and the number of complex %\av{what is a complex punctuation mark?}\anote{answered}
punctuation marks (i.e.\ punctuation within the target word/MWE, such as hyphens, that could denote added  complexity). %\nando{why would there be commas or semi-colons within a target word/MWE? I understand the hyphens, but not the other punctuation marks.}\anote{addressed}
We would expect certain affixes to be useful features, as language users use sub-word particles like these to identify unknown words: by breaking up a word like ``granted'' into ``grant-'' and ``-ed'', readers can fall back on their knowledge of these component pieces to clarify the whole. A total of 9 sub-word features were used in the monolingual models.

% Our model found that Prefixes like "ex-" or suffixes like "-ed" were correlated with lower complexity.\av{Avoid talking about the results, they should be left for the experiments subsection. Here give the linguistic sense}

Finally, \textbf{sentence level} features with linguistic motivations were also considered. 
Long sentences could be harder to understand, which makes it more difficult to figure out the meaning of unknown words contained within them. 
Also, long sentences are more likely to include more unknown words or ambiguous references.
Therefore, we considered sentence length (i.e., number of tokens in the sentence) as a feature. 
In addition, we extracted N-grams (unigrams, bigrams and trigrams) from the whole sentence, since certain sentence constructions can help a reader understand the target word/MWE. For example, ``A of the B'' suggests a relation between A and B.
We used 2 sentence-level features in total. 

%Details on the resources used for extracting each of the 24 features can be found in the Supplementary Material.
%were important for some datasets. For example, the bigram ``of the'' was correlated with lower complexity. These kinds of simple constructions are likely to help readers to understand the words that appear elsewhere in the sentence.

%AV: Don't talk about the experiments yet!
%The most influential feature types were those relating to word function and sub-word features.

% Maybe something about the fact that many other papers approached the problem by looking for large external corpora, whereas we achieved relatively high performance with relatively few corpus-based features?

%\subsection{Datasets}

%AV: Moved it here to be on the same page as the results
\subsection{Experiments and Results}

Following \citet{sharedtask:yimam-etal:2018}, we used Macro-F1 score to evaluate performance and for comparison with previous work on the datasets. 
We used Logistic Regression for all our experiments, as it allowed for easy exploration of feature combinations, and in initial experiments we found that it performed better than Random Forests. 
We evaluated both using the full feature set described before, as well as a two-feature baseline using the number of tokens of the target and its language-normalised  number of  characters.
Results of our monolingual experiments are shown in Table~\ref{table:mono_results}. 

\begin{table}[hbt]
\small
\centering
\begin{tabular}{@{}lrrrrrr@{}}
\toprule
Dataset             & \multicolumn{2}{c}{Dev}  & \phantom{l} & \multicolumn{3}{c}{Test}\\
\cmidrule{2-3} \cmidrule{5-7}
                    & BL    & MA    & & BL   & MA  & SotA \\
\midrule
EN - News         & 83.6  & 85.5  & & 69.7  & 86.0  & \bf87.4         \\
EN - WikiNews     & 80.4  & 82.8  & & 65.8  & 81.6  & \bf 84.0         \\
EN - Wikipedia    & 74.2  & 76.6  & & 70.1  & 76.1  & \bf 81.2         \\
ES              & 78.0  & 77.1  & & 69.6  & \bf 77.6 & 77.0         \\
DE              & 79.5  & 74.6  & & 72.4  & 74.8 & \bf 75.5       \\ 
\midrule
Mean            & 79.1  & 79.3  & & 69.5  & 79.2 & {\color{NACol}N/A}     \\ 
\bottomrule
\end{tabular}
\caption{Macro-F1 for the baseline (BL), our monolingual approach (MA), and the state of the art (SotA) on the Dev and Test splits of each dataset.}
\label{table:mono_results}
\end{table}

In the test set, our baseline results (BL in Table~\ref{table:mono_results}) are strong, especially in German. 
%\nando{Why German? Aren't they more impressive in Spanish where we beat the SotA?} %AV: it talks about the baseline (BL), not the full model (MA)
Our full 25-features model improves on the baseline in all cases, with the biggest increase of over 16 percentage points seen for the EN-News dataset. 
Our system beats the best performing system from the Shared Task in Spanish (77.0) and German (74.5), both obtained by \citet{cwi:kajiwara-komachi:2018}. 
However, the state of the art for German remains the Shared Task baseline (75.5) \citep{sharedtask:yimam-etal:2018}. 
The best results for all three English datasets were obtained by \citet{cwi:gooding-kochmar:2018}; ours is within two percentage points of their News dataset score. 
Furthermore, the mean score for our system (79.2) is close to the mean of the best performing models (81.0), which are different systems, while using simpler features and learning algorithm. 
The best-performing model in English, for example, used Adaboost with 5000 estimators \citep{cwi:gooding-kochmar:2018}.

% \av{That's Gooding right? mention it explicitly}
% which didn't require hyper-parameter tuning.%\av{can we give some evidence for this?} 
%\nando{for ``simpler'' we could mention the features and learning algorithm, but I don't know about ``quick''. AV: Yes, let's do this, quick is hard to tell unless you run everything yourself}

%Feature importance

\section{Cross-lingual Models} % Aneeq, Elizabeth, Pierre
\label{sec:crosslingual}

\subsection{Features Description} 
%\nando{I'm tempted to change the name of this section to Feature Selection or Language-Independent Features, and add the first paragraph of 3.2, which is related to the features}
Linguistically, the cross-lingual approach can be motivated by the relation between certain languages (such as French and Spanish both being Romance languages). In addition, there may be features identifying complex words that are shared even across language families.

%Main idea: train the model using a language (or set of languages) that is different from the language (or set of languages) on which the model is tested.

%Motivation behind main idea: much data in some languages, less in others, there might be some features that nearly all people find complex, languages with common origins might be usable with each other.

%\subsection{Features}
To be able to test a model on a language that was unseen during training, the features the model works with must be cross-lingual (or language-independent) themselves. For example, the words themselves are unlikely to transfer across languages (apart from those that happen to be spelled identically), but the popularity of the words would transfer.
%bag of words obtained from English training data will not be applicable to French, whereas certain suffixes like ``-able'' are common across both of those languages. % Features used must be applicab
% they must be applicable to all languages and cannot rely upon a specific language to work.\av{tautology}
This rules out some of the features we used for the monolingual approach (see Sec.~\ref{sec:mono_features}), as they were language-dependent. 
One such feature is N-grams for the target word/MWE, %and surrounding context 
%\anote{removed reference to context}
which depend on the language, and so will only occur with extreme sparsity outside of their source language. For example, if applying a system trained on English to unseen French, the English phrases ``à la mode'' or ``film noir'' might reoccur in the French, since they originate from that language, but these are rare exceptions. What is more, a French loan-phrase may have different complexity characteristics to the same N-grams occurring in their native language. Therefore, we did not use these features in the cross-lingual system. %\nando{which features of the 24 were ruled out? AV: or some examples at least} \pmfin{Beefed this up a bit, feel free to cut}
%\av{expected to see here which features are left to be considered in our experiments}
%AV: remove references to results
%To analyse which of the remaining features candidates would aid the cross-lingual approach, we performed an iterative analysis, the results of which are described in Section \ref{results}.

% NOTE(Pierre): These are all of the features that our crosslingual system could have used:
% is_nounphrase
% len_tokens_norm
% hypernym_count
% len_chars_norm
% len_tokens
% len_syllables
% consonant_freq
% gr_or_lat
% is_capitalised
% num_complex_punct
% averaged_chars_per_word
% sent_length
% unigram_prob
% char_n_gram_feats
% sent_n_gram_feats
% iob_tags
% lemma_feats
% bag_of_shapes
% pos_tag_counts
% NER_tag_counts 

% NOTE(Pierre): These are all of the features that our crosslingual system ended up using:
% len_syllables
% len_tokens
% num_complex_punct
% sent_length
% unigram_prob

%spacy citation \citep{spacy2} AV: Ican't figure out why it breaks the latex making

\subsection{Experiments and Results} % FA: For lack of a better name

To find out which features were best suited for the cross-lingual approach, we performed an iterative ablation analysis %\nando{Did the ablation study started with the 24 features or with only the language-independent ones? If the latter, how many were there?} \pmfin{I believe it  started just with len tokens, then features were added and removed (from the set of 20) iteratively until we had 6}
(see Appendix~\ref{appe:crosslingual_ablation} for details). %\anote{added ref to table 9}
Using this process, we arrived at our final cross-lingual feature set: number of syllables in the target, number of tokens in the target, number of complex punctuation marks (such as hyphens), sentence length, and unigram probabilities. %\av{what is complex punctuation?}\anote{gave example}

Furthermore, we analyse the effect of different language combinations on the performance of the cross-lingual model in order to investigate how the relationship between the languages trained and tested on would influence model performance. 
Recall that we only have training data for English, Spanish and German, but not French. We train models using all possible combinations (each language independently, each pairing, and all three) and evaluate on each of the four languages that have test data (i.e.\ the former three and French), excluding training combinations that include the test language. 
Results are shown in Table \ref{table:language-ablation}.
%We trained the model using only one of these languages, and then using all possible combinations of two languages and finally all three languages together. When testing on a particular language, we only considered those training language combinations that did not include that language. 

% \begin{figure}[ht]
%     \includegraphics[width=0.45\textwidth]{TestNotSeenInTrainBar.png}
%     \caption{Effect of different combinations of training languages on Macro-F1 score. English is averaged over the 3 datasets, with error-bars.}
%     \label{fig:language-ablation}
% \end{figure}

\begin{table}[!htb]

\centering \small
\begin{tabular}{@{}cccllrr@{}}
\toprule
EN & ES & DE & Eval & Source & Test  & Dev   \\
\midrule
							& \cellcolor{EsCol}\checkmark	& \cellcolor{DeCol}\checkmark	& \cellcolor{EnCol}EN	& WikiNews	& 61.8	& 63.7 \\
					
							&								& \cellcolor{DeCol}\checkmark	& \cellcolor{EnCol}EN	& WikiNews   & 62.3 & 63.6 \\
         
							& \cellcolor{EsCol}\checkmark	&								&  \cellcolor{EnCol}EN	& WikiNews   & 61.6 & 63.8 \\
         
        %\midrule
        
							& \cellcolor{EsCol}\checkmark	& \cellcolor{DeCol}\checkmark	& \cellcolor{EnCol}EN	& Wikipedia  & 62.8 & 64.4 \\
         
							&								& \cellcolor{DeCol}\checkmark	&  \cellcolor{EnCol}EN	& Wikipedia  & 62.6 & 64.4 \\
         
							& \cellcolor{EsCol}\checkmark	&								&  \cellcolor{EnCol}EN	& Wikipedia  & 63.1 & 65.2 \\
         
        %\midrule
        
							& \cellcolor{EsCol}\checkmark	& \cellcolor{DeCol}\checkmark	& \cellcolor{EnCol}EN	& News       & 67.1 & 65.6 \\
         
							&								& \cellcolor{DeCol}\checkmark	&  \cellcolor{EnCol}EN	& News       & 67.0  & 65.6 \\
         
							& \cellcolor{EsCol}\checkmark	&								&  \cellcolor{EnCol}EN	& News       & \textbf{67.2} & \textbf{65.9} \\
         
         \midrule
         
\cellcolor{EnCol}\checkmark	&								& \cellcolor{DeCol}\checkmark	&  \cellcolor{EsCol}ES	& {\color{NACol}N/A}	& 70.8 & 71.3 \\

							&								& \cellcolor{DeCol}\checkmark	&  \cellcolor{EsCol}ES	& {\color{NACol}N/A}	& \textbf{72.6} & \textbf{74.1} \\
         
\cellcolor{EnCol}\checkmark	&								&								&  \cellcolor{EsCol}ES	& {\color{NACol}N/A}	& 69.1 & 70.0   \\

\midrule

\cellcolor{EnCol}\checkmark	& \cellcolor{EsCol}\checkmark	&								& \cellcolor{DeCol}DE	& {\color{NACol}N/A}	& \textbf{73.4} & \textbf{78.3} \\

							& \cellcolor{EsCol}\checkmark	&								&  \cellcolor{DeCol}DE	& {\color{NACol}N/A}	& 72.6 & 77.4 \\
         
\cellcolor{EnCol}\checkmark	&								&								& \cellcolor{DeCol}DE	& {\color{NACol}N/A}	& 73.0  & 76.0  \\

\midrule

\cellcolor{EnCol}\checkmark	& \cellcolor{EsCol}\checkmark	& \cellcolor{DeCol}\checkmark	& \cellcolor{FrCol}FR	& {\color{NACol}N/A}	& 73.1 & {\color{NACol}N/A}      \\

							& \cellcolor{EsCol}\checkmark	& \cellcolor{DeCol}\checkmark	& \cellcolor{FrCol}FR	& {\color{NACol}N/A}	& 75.7 & {\color{NACol}N/A}  \\
         
\cellcolor{EnCol}\checkmark	& \cellcolor{EsCol}\checkmark	&								& \cellcolor{FrCol}FR	& {\color{NACol}N/A}	& 73.4 & {\color{NACol}N/A}   \\

\cellcolor{EnCol}\checkmark	&								& \cellcolor{DeCol}\checkmark	& \cellcolor{FrCol}FR	& {\color{NACol}N/A}	& 70.5 & {\color{NACol}N/A}   \\

							&								& \cellcolor{DeCol}\checkmark	& \cellcolor{FrCol}FR	& {\color{NACol}N/A}	& \textbf{75.8} & {\color{NACol}N/A}   \\
         
							& \cellcolor{EsCol}\checkmark	&								& \cellcolor{FrCol}FR	& {\color{NACol}N/A}	& 73.4 & {\color{NACol}N/A}   \\
         
\cellcolor{EnCol}\checkmark	&								&								& \cellcolor{FrCol}FR	& {\color{NACol}N/A}	& 69.2 & {\color{NACol}N/A}  \\

\bottomrule
\end{tabular}
\caption{Comparison of Test and Dev results for all permutations of training languages.}
\label{table:language-ablation}
\end{table}

%\pmfin{The Table is still a bit wide, but it's hard to cut down. Also, should it feature colours to make it a bit clearer?}
%\nando{I made the table fit in the column by changing the size of the font. What's your idea of using colours? How would they help?}
% \pmfin{I've added the initial colours to show the different permutations more clearly. I'm also considering adding a gradient for the numbers, to show their relative scale. (though this would take a reasonable amount of effort, so I haven't done it yet)}
% \nando{I don't think that's necessary for the paper. For the poster, that could look pretty nice!}

When testing on French, we achieved the highest performance by training on German only (75.8), followed closely by training on a combination of German and Spanish (75.7) and only Spanish (75.5). %This is remarkable as there is no training data for French and our model performance is very close to the state of the art model for French as shown in Table \ref{table:cross-lingual results}.
%\av{I don't get this: no other system has training data for French, what's remarkable? Maybe remove the sentence altogether?}\anote{I don't get this sentence either so I've commented it out}
The worst performance was achieved by training only on English (69.2), and the performance also noticeably decreased for all training combinations that included English.

When testing on German, language choice had a weaker effect. 
The highest score came from combining English and Spanish (73.4), but using only one of those languages gave comparable results (72.6 for Spanish, 73.0 for English).

For Spanish, the best results were achieved when training only on German (72.6). 
Adding English to the training languages decreased the performance (70.8), which was even lower when training only on English (69.1).

% Not sure whether to use this one instead?
% \begin{figure}[h]
% \includegraphics[width=\textwidth/2]{TestSeenInTrainBar.png}
% \end{figure}

%What is noteworthy about these observations is
It is noteworthy that adding English to the training languages noticeably decreases performance for both Spanish and French, but not for German. One possible reason for Spanish and French not benefiting from English when German does is that both English and German are Germanic languages, whereas Spanish and French are Romance languages. 
%\av{Makes sense! Can we support it by looking at the datasets to see what are the CWI features that these language families share?} 
Another possible explanation for the decrease of performance caused by training with English is that there are inconsistencies in the way MWEs in the datasets were labelled across languages, %\av{clarify: inconsistencies across languages. AS: done}, 
which we explore in Sec.~\ref{dataset}.

\begin{table}[t]
\small
\centering
\begin{tabular}{@{}lrrr@{}}
\toprule
                    &  Spanish &  German  &  French \\
\midrule
Monolingual SotA    & \bf 77.0          & \bf 75.5          & {\color{NACol}N/A} \\
Cross-lingual SotA  & {\color{NACol}N/A}& {\color{NACol}N/A}& \bf 76.0 \\
Our cross-lingual   & 72.6              & 73.4              & 75.8 \\
\bottomrule
\end{tabular}
\caption{Comparison between the monolingual and cross-lingual state of the art (SotA), and our cross-lingual system.}
\label{table:cross-lingual results}
\end{table}

We finally compare our cross-lingual models against the state of the art: the best monolingual system for Spanish and German, and the best cross-lingual system for French, where no monolingual systems exist.
As Table \ref{table:cross-lingual results} shows, our cross-lingual models come close to the best monolingual models for Spanish and especially for German. 
This is remarkable given how simple our model and features are, and that the approaches we compare against
%. We seem to be able to train a model using only a few simple language-independent features and then get results on a previously unseen language that are nearly as good as the results achieved by specifically 
train complex models for each language. 
Furthermore, this points towards the possibility of extending CWI to more languages which lack training data. %using this approach. 

Finally, Table \ref{table:romance-germanic} %\av{on the table testing language is irrelevant right?} 
compares the coefficients for models trained on Romance and Germanic languages. Notably, use of complex punctuation (such as the hyphenation in ``laser-activated'' or ``drug-related'') %\av{what is complex punctuation! Would be good to give examples here} \pmfin{done} 
and the number of tokens are inversely correlated w.r.t. the word or MWE being complex. More words in the target was correlated with complexity for English and German, and inversely correlated for Spanish.

% We can remove the first nine rows of this table if we need more space.

% \begin{table}[t]
% \small
% \centering
% \begin{tabular}{@{}llll@{}}
% \toprule
% Feature         & Train & Test  & Coefficient \\ 
% \midrule
% len\_syllables  & EN    & DE    & 0.620 \\
% len\_syllables  & DE    & EN    & 0.630 \\
% len\_syllables  & ES    & FR    & 0.624 \\
% \midrule
% sent\_length    & EN    & DE    & -0.006 \\ 
% sent\_length    & DE    & EN    & -0.010 \\ 
% sent\_length    & ES    & FR    & -0.006 \\ 
% \midrule
% unigram\_prob   & EN    & DE    & -0.230 \\ 
% unigram\_prob   & DE    & EN    & -0.372 \\ 
% unigram\_prob   & ES    & FR    & -0.456 \\ 
% \midrule
% \textbf{num\_complex\_punct }   & \textbf{EN}    & \textbf{DE}    & \textbf{-0.693} \\ % These last two are the most significant differences.
% \textbf{num\_complex\_punct }   & \textbf{DE}    & \textbf{EN}    & \textbf{-0.559} \\ 
% \textbf{num\_complex\_punct }   & \textbf{ES}    & \textbf{FR} & \textbf{1.111} \\ 
% \midrule
% \textbf{len\_tokens}     & \textbf{EN}    & \textbf{DE}    & \textbf{-2.200} \\ 
% \textbf{len\_tokens}     & \textbf{DE}    & \textbf{EN}    & \textbf{-0.534} \\ 
% \textbf{len\_tokens}     & \textbf{ES}    & \textbf{FR}   & \textbf{1.420} \\ 
% \bottomrule
% \end{tabular}
% \caption{Coefficients for cross-lingual models trained on Germanic and Romance languages.} 
% \label{table:romance-germanic}
% \end{table}

\begin{table}[htb]
\small
\centering
\begin{tabular}{@{}p{0.62\columnwidth}lr@{}}
\toprule
Feature                                                 & Train & Coefficient \\ 
\midrule
\multirow{3}{*}{number of complex punctuation marks}    & EN & -0.693 \\ % These last two are the most significant differences.
                                                        & DE & -0.559 \\ 
                                                        & ES & 1.111 \\ 
\midrule
\multirow{3}{*}{number of tokens}                       & EN    & -2.200 \\ 
                                                        & DE    & -0.534 \\ 
                                                        & ES    & 1.420 \\ 
\bottomrule
\end{tabular}
\caption{Coefficients for cross-lingual models trained on Germanic and Romance languages.} 
\label{table:romance-germanic}
\end{table}

% \begin{table}[t]
% \small
% \centering
% \begin{tabular}{@{}llll@{}}
% \toprule
% Feature         & Train & Test  & Coefficient \\ 
% \midrule
% num\_complex\_punct   & EN    & DE    & -0.693 \\ % These last two are the most significant differences.
% num\_complex\_punct    & DE    & EN    & -0.559 \\ 
% num\_complex\_punct    & ES    & FR & 1.111 \\ 
% \midrule
% len\_tokens     & EN    & DE    & -2.200 \\ 
% len\_tokens     & DE    & EN    & -0.534 \\ 
% len\_tokens     & ES    & FR   & 1.420 \\ 
% \bottomrule
% \end{tabular}
% \caption{Coefficients for cross-lingual models trained on Germanic and Romance languages.} 
% \label{table:romance-germanic}
% \end{table}

\section{Dataset Analysis}
\label{dataset}

%\subsection{Ablation Study}

%\subsection{Error Analysis}
% Alison
While examining our models' incorrect predictions, we observed inconsistencies in labelling in the datasets between target MWEs and their sub-words/sub-expressions (SWs). %\av{maybe reuse the terminology when describing the features?}
%
%Data annotation for CWI is a difficult task given its subjective nature. %\av{Maybe we should start this section by motivating why we do this analysis and then proceed with this. as: added}

The First CWI Shared Task~\citep{sharedtask:paetzold-specia:2016} used the annotations of a group (i.e.\ ten annotators on the training data) to predict the annotation of an individual (i.e.\ one annotator on the test data). %\nando{This is not clear. If I remember correctly, what happened was that the test set was annotated by only one person, while the training set had more? AV: FA could you add this point?}\anote{I agree the test set was annotated by a single person, and the training set by 10, I've rewritten the sentence to try and make that point clearer}
The resulting inconsistencies in labelling may have contributed to the low F-scores of systems in the task \citep{cwi:zampieri-etal:2017}.
%That study found that half of the training words labelled complex by at least half of the annotators, and that also appeared in the test data, were given the opposite label in the test data. \av{can remove this sentence if needed for space, it is not our work (and I found it a bit hard to follow)}
%\nando{It could be interesting to mention some of the inconsistencies they found in that dataset, and then contrast them to the ones found in this new and ``better'' dataset. AS: added} 
Although the Second CWI Shared Task improved on the first by having multiple annotators for all splits of the data, %\nando{for the test set}\anote{clarified}, 
it contains some labelling inconsistencies arising from annotators now being able to  label phrases, and words within them, separately.

More concretely, we found that across all datasets, 72\% of  MWEs contain at least one SW with the opposite label (see Table~\ref{table:data_noise}). While this makes sense in some cases, every SW in 25\% of MWE instances has the opposite label. For example, ``numerous falsifications and ballot stuffing" is not annotated as complex, despite
its SWs ``numerous", ``numerous falsifications", ``falsifications'', ``ballot'', ``ballot stuffing" and ``stuffing" all being complex. Conversely, ``crise des march\'es du cr\'edit" is complex, despite ``crise", ``march\'es" and ``cr\'edit" being labelled non-complex. It is difficult to see how classifiers that extract features for MWEs from their individual SWs could predict the labels of both correctly. %\av{So what does this mean for other people building systems? What percentage of instances were affected? AS: added}

Furthermore, every target MWE in the Spanish, German and French datasets is labelled complex. %\av{doesn't this explain why adding English decreases the performance always? Good to know. AS: added comment}
This may bias a classifier trained on the Spanish or German datasets towards learning MWEs and long individual words (if length is a feature) are complex. 
In particular, this observation may help explain why adding English as a training language decreased the performance of our cross-lingual system when testing on French and Spanish (where all MWEs are complex). 
An analysis in \citet{multitask:bingel-bjerva:2018} further found %French complex words/MWEs tended to be longer, and 
that their cross-lingual French model was effective at predicting long complex words/MWEs but had difficulty %correctly 
predicting long non-complex words.%\av{Does long words is the same as MWE? If so stick with MWE. AS: no, long means a high number of characters in the word/MWE; I've tried to make that clearer.}.

%Given the level of noise in the data, it is unlikely a reasonable classifier could make the correct prediction in many cases. 
%This level of noise makes correct predictions very difficult in some cases.  

%\nando{Bingel and Bjerva also analyse the data, but with respect to the length of the complex MWE. Maybe it's worth mentioning it to complement our findings. AS: added}

\begin{table}[t]
\small
\centering
\begin{tabular}{@{}lrrrr@{}}
\toprule
        &  C &  NC  &  $\geq$ 1 Irreg. &  All Irreg. \\
\midrule
English &   3,750 &     982 &                3,315 &             950 \\
Spanish &   2,309 &       0 &                1,747 &             760 \\
German  &    502 &       0 &                 374 &             178 \\
French  &    242 &       0 &                 192 &              82 \\
\midrule
Total   &   6,803 &     982 &                5,628 &            1,970 \\
\bottomrule
\end{tabular}
\caption{MWE annotation analysis: numbers of MWEs labelled complex (C) and non-complex (NC), numbers with at least one SW ($\geq 1$ Irreg) and all SWs (All Irreg.) having the opposite label.}
\label{table:data_noise}
\end{table}

It is also worth noting that considering a word or MWE as complex is subjective and may differ from person to person, even within the same target audience. 
\citet{bingel2018predicting} investigated predicting complex words based on the gaze patterns of children with reading difficulties. 
They found a high degree of specificity in misreadings between children, that is, which words they found complex when reading aloud. 
This variety of complexity judgements even within one target group points towards the high degree of subjectivity in the task, which may also partly explain the inconsistencies in the dataset.

\section{Conclusion and Future Work}
The monolingual and cross-lingual models presented achieve comparable results against more complex, language-specific state-of-the-art models, and thus can serve as strong baselines for future research in CWI. 
%\av{remind the reader why these are good baselines}
In addition, our analysis of the dataset could help in the design of better guidelines when crowdsourcing annotations for the task. 
Dataset creators may wish to only allow single words to be chosen as complex to avoid labelling inconsistencies. 
In case MWEs are being permitted, we suggest instructing annotators to chose the smallest part of a phrase they find complex (French annotators for the Second CWI Shared Task sometimes grouped individual complex words into a complex MWE \citep{sharedtask:yimam-etal:2018}).

\section*{Acknowledgements}
This work was initiated in a class project for the NLP module at the University of Sheffield. The authors would like to acknowledge the contributions of Thomas Dakin, Sanjana Khot and Harry Wells  who contributed their project code to this work.
Andreas Vlachos is supported by the EPSRC grant eNeMILP (EP/R021643/1).

\bibliography{naaclhlt2019}
\bibliographystyle{acl_natbib}

\clearpage
\appendix

\onecolumn
\section{Detailed Feature Set}
\label{appe:detailed_features}

\begin{table}[htb]
\centering \small
\ra{1.3}
\begin{tabular}{@{}llp{0.3\textwidth}p{0.36\textwidth}@{}}
\toprule
Level & Name                 &  Description                             & Resource \\ 
\midrule
\multirow{22}{0.1\textwidth}{Target\newline word/MWE} & NER\_tag\_counts           & Counts of each Named Entity tag in target & spaCy \\

& pos\_tag\_counts           & Counts of each part-of-speech tag in target            & spaCy \\

& hypernym\_count            & Number of hypernyms       & WordNet (NLTK) \\

& len\_tokens                & Absolute length in tokens   & N/A \\

& len\_tokens\_norm          & Normalised length in tokens   & N/A \\

& len\_chars\_norm           & Normalised length in characters & N/A \\

& unigram\_prob              & Log of the product of unigram probabilities   & EN: Brown Corpus (NLTK)\newline ES: CESS-ESP (NLTK)\newline DE: TIGER Corpus\tablefootnote{\url{https://www.ims.uni-stuttgart.de/forschung/ressourcen/korpora/tiger.html}} \newline FR: Europal\tablefootnote{\url{http://www.statmt.org/europarl/}} \\

& bag\_of\_shapes            & Bag of morphological shapes          & spaCy \\

& rare\_word\_count          & Count of rare words in target    & EN: subset of Google's Trillion Word Corpus\tablefootnote{\url{https://github.com/first20hours/google-10000-english}} \newline DE: list of the most common 3,000 words\tablefootnote{\url{http://germanvocab.com/}}\newline ES: word frequency list by  M. Buchmeier\tablefootnote{\url{https://en.wiktionary.org/wiki/User:Matthias_Buchmeier/Spanish_frequency_list-1-5000}}\\

& rare\_trigram\_count       & Count of rare trigrams in target    &     Same as rare\_word\_count\\

& is\_stop                   & Frequency of stopwords in target    &     NLTK, Ranks NL\tablefootnote{\url{https://www.ranks.nl/stopwords}} \\
  
& is\_nounphrase             & If target is a noun phrase  & spaCy \\

& avg\_chars\_per\_word       & Avg. word length (in characters) of the target  & N/A \\

& iob\_tags                  & Count of BIO tags in target  & spaCy  \\

\midrule

\multirow{14}{*}{Sub-word}                     

& lemma\_feats               & Bag of lemmas for target sentence    & spaCy \\

& len\_sylls\                 & Length of target in syllables & Pyphen\tablefootnote{\url{https://pyphen.org/}} \\

& num\_complex\_punct        & Count of complex punctuation in target  & N/A \\

& char\_n\_gram\_feats       & Character N-Grams, incl. prefixes and suffixes              & N/A \\

& char\_tri\_sum             & Sum of character trigrams' corpus frequencies   & EN: Brown Corpus (NLTK)\newline ES: CESS-ESP (NLTK)\newline DE: TIGER Corpus \\

& char\_tri\_avg             & Average of character trigrams' corpus frequencies & same as char\_tri\_sum  \\

& consonant\_freq            & Count of consonants in target   & N/A \\

& gr\_or\_lat                & If target has Greek or Latin affixes & List of Greek and Latin roots in English\tablefootnote{\url{https://www.oakton.edu/user/3/gherrera/Greek\%20and\%20Latin\%20Roots\%20in\%20English/greek_and_latin_roots.pdf}}\\

& is\_capitalised            & If target's first letter is uppercased  & N/A \\

\midrule
\multirow{3}{*}{Sentence}  & sent\_length           & Number of tokens in the sentence       & N/A \\
                           & sent\_n\_gram\_feats   & Unigrams, bigrams and trigrams in the sentence       & N/A \\
\bottomrule                                             
\end{tabular}
\caption{Monolingual and Cross-lingual Feature Set Summary}
\label{table:mono_features}
\end{table}

\clearpage
\section{Cross-lingual Features Ablation}
\label{appe:crosslingual_ablation}

\begin{table}[hbt]
\centering \small \ra{1.2}
\begin{tabular}{@{}cp{0.25\columnwidth}p{0.26\columnwidth}p{0.26\columnwidth}@{}}
\toprule
 Iteration & Current features      & Features increasing performance & Features decreasing performance \\ \midrule

\multirow{6}{*}{1} & len\_tokens &  num\_complex\_punct\newline len\_sylls\newline sent\_length  & is\_nounphrase\newline len\_tokens\_norm\newline consonant\_freq\newline is\_capitalised\newline bag\_of\_shapes\newline pos\_tag\_count \\ 
 \midrule
 
\multirow{4}{*}{2} & len\_tokens\newline len\_sylls\newline num\_complex\_punct\newline sent\_length &    unigram\_prob    & gr\_or\_lat     \\ \midrule
 
\multirow{5}{*}{3} & len\_tokens\newline len\_sylls\newline num\_complex\_punct\newline sent\_length\newline unigram\_prob &    & char\_ngram\_feats\newline iob\_tags\newline lemma\_feats\newline NER\_tag\_counts         \\ 
 \bottomrule
\end{tabular}
\caption{Ablation analysis for the cross-lingual features }
\label{table:cross_ling_features_full}
\end{table}

\end{document}